\documentclass[letterpaper]{llncs}

\usepackage{amsmath,amssymb}
\usepackage{graphicx}
\usepackage{subfigure}
\usepackage{multirow}
\usepackage{booktabs}
\usepackage{times}
\usepackage[width=122mm,left=12mm,paperwidth=146mm,height=193mm,top=12mm,paperheight=217mm]{geometry}
\usepackage[pagebackref=true,colorlinks,citecolor=red,bookmarks=false]{hyperref}

\begin{document}

\pagestyle{headings} 

\title{Spatio-Temporal Tuples Transformer for Skeleton-Based Action Recognition}
\author{Helei Qiu, Biao Hou*, Bo Ren, Xiaohua Zhang}
\institute{Xidian University}

\maketitle

\begin{abstract}

\small
Capturing the dependencies between joints is critical in skeleton-based action recognition task. Transformer shows great potential to model the correlation of important joints. However, the existing Transformer-based methods cannot capture the correlation of different joints between frames, which the correlation is very useful since different body parts (such as the arms and legs in "long jump") between adjacent frames move together. Focus on this problem, A novel spatio-temporal tuples Transformer (STTFormer) method is proposed. The skeleton sequence is divided into several parts, and several consecutive frames contained in each part are encoded. And then a spatio-temporal tuples self-attention module is proposed to capture the relationship of different joints in consecutive frames. In addition, a feature aggregation module is introduced between non-adjacent frames to enhance the ability to distinguish similar actions. Compared with the state-of-the-art methods, our method achieves better performance on two large-scale datasets.

\end{abstract}

\section{Introduction}

Recently, skeleton-based action recognition has attracted substantial attention since its compact skeleton data representation makes the models more efficient and robust to complex background, illumination conditions, viewpoint changes and other environmental noise. In addition, the development of cost-effective depth cameras and human pose estimation methods makes it easier to obtain human skeleton information.

The raw skeleton data is usually converted into point sequences, pseudo images or graph, and then input into deep network (such as recurrent neural network (RNN), convolutional neural network (CNN) or graph neural network (GCN)) for feature extraction. However, the RNN-based methods \cite{RNN_GRU_2017,RNN_LSTM_20171,RNN_LSTM_20172} processes sequences recursively, and the CNN-based methods \cite{CNN_20171,CNN_20172,CNN_2020,CNN_2021} performs local operations on a fixed size window. Both methods capture only short-range correlations. The GCN-based methods \cite{ST-GCN,Two-Stream,MS-G3D,Tripool,Sybio-GNN,PoseC3D,Hyper-GNN} rely on the inherent graph topology of the human body, and cannot effectively use the correlation between unconnected joints (such as hands and feet in “put on shoes”). In general, the above methods cannot effectively model the long-term dependence of sequences and the global correlation of spatio-temporal joints.

\begin{figure}
  \centering
  \subfigure[]{\label{Skeleton1}\includegraphics[scale=0.8]{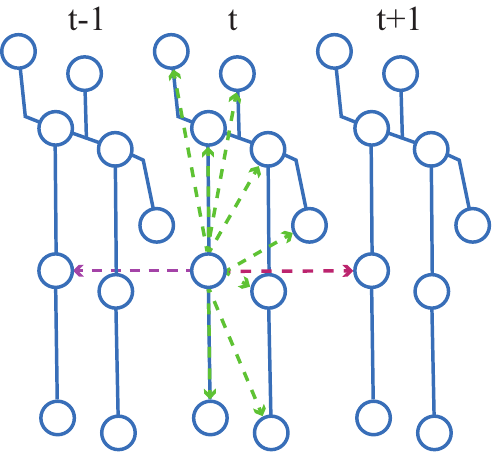}}
  \hspace{2mm}
  \subfigure[]{\label{Skeleton2}\includegraphics[scale=0.8]{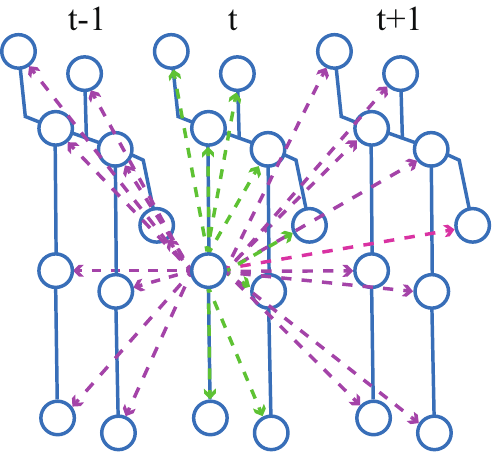}}
  \caption{Two spatio-temporal self-attention schemes. (a): This scheme only establishes the relationship of intra-frame joints and the same joints between inter-frames. (b): This scheme captures the relationship of all joints in several consecutive frames at the same time.}

  \label{Skeleton}
\end{figure}

The Transformer-based methods does not depend on the human structure, and can model the relationship between all joints compared with the above-mentioned methods.
Considering this advantage, Transformer \cite{Transformer} is applied to skeleton-based action recognition tasks \cite{Non-Local,ST-TR,DSTANet}.  How to use Transformer to model spatio-temporal correlation is crucial since the spatiotemporal joints of skeleton data are correlated. \cite{Non-Local} regards spatio-temporal skeleton data as a single sequence to capture the related information of all joints. However, this strategy is unreasonable since the spatial and temporal joints have different semantic information. Specifically, the relationship between spatial joints reflects the interaction between various parts of the human body, and the same joints between temporal frames represent the motion trajectory of a certain part of the human body. In addition, this method needs to calculate the self-attention of all joints in the sequence at the same time, which will significantly increase the computational cost. Focus on these problems, \cite{ST-TR} introduces self-attention into graph convolution, and uses a spatial and temporal self-attention module to model the correlation of intra-frame and inter-frame joints respectively. Similarly, \cite{DSTANet} uses pure self-attention to capture the relationship of the spatio-temporal joints. However, these methods only focus on the same joints between frames (Fig. \ref{Skeleton1}), and the extracted related motion features are too simple.
It is observed that different joints in several consecutive frames are related. For example, in the action "long jump", the arms of the previous frame are related to the legs of the next frame since the joints of these parts move together in cases of action movement. Therefore, it is very useful to extract the related features of different joints between adjacent frames.

Based on the above observations, we construct a novel spatio-temporal tuples Transformer (STTFormer) model for skeleton-based action recognition. Specifically, a skeleton sequence is divided into several non-overlapping parts. Each part is called a "tuple" and contains several consecutive frames. Because different joints in several consecutive frames have correlation, each tuple is flattened into a short sequence, and then a spatio-temporal tuple self-attention (STTA) module is used to extract the related features of joints in each short sequence simply and effectively as shown in Fig. \ref{Skeleton2}. This method not only capture the correlation of different joints between consecutive frames, but also hardly increase the computational cost. Because although all joints of several frames need to be modeled at the same time, the size of time dimension is greatly reduced. In addition, we aggregate features on the time dimension composed of several tuples. If a tuple is regarded as a sub-action, the inter-frame feature aggregation (IFFA) can be regarded as the integration of a series of sub-actions. the IFFA module will help distinguish similar actions. Finally, multi-mode data is also used to further improve performance. The code will be made publicly available at \url{https://github.com/heleiqiu/STTFormer}

The main contributions of this work are as follows:
\begin{itemize}
  \item A spatio-temporal tuple encoding strategy is proposed to explicitly flatten the joints of several consecutive frames, so that our model can capture the related information of different joints between frames.
  \item A spatio-temporal tuple Transformer is proposed, in which the spatio-temporal tuple attention is used to capture the related features of the joints in each tuple, and the inter-frame feature aggregation module is used to integrate all tuples.
  \item We conducted extensive experiments on two challenging benchmarks. Ablation experiments verify the effectiveness of each component of our model, and the performance of our method exceeds the existing state-of-the-art methods.
\end{itemize}

\section{Related Work}

\subsection{Self-Attention Mechanism}
Recently, Transformer \cite{Transformer} has become the leading language model in natural language processing. The self-attention is an important component of Transformer, which can learn the relationships between each element of a sequence. Transformer can handle very long sequences and solves the problem that LSTM and RNN networks cannot effectively model long-term sequences. In addition, the multi-headed self-attention mechanism can process sentences in parallel, rather than processing sentences word by word recursively like LSTM and RNN networks. 

Due to the advantages of self-attention, which has also been introduced into computer vision tasks such as image classification and recognition \cite{CvT,ViT}, object detection \cite{DETR,ACT} and action recognition \cite{TimeSformer,ST-TR,ViViT}. \cite{CvT} combined CNN and self-attention to model the local and global dependencies for image classification. \cite{TimeSformer} used self-attention to learn spatio-temporal features from a sequence of frame-level patches for video action recognition. Based on GCN, \cite{ST-TR} used self-attention instead of regular graph convolution in space and time for skeleton-based action recognition. 
In contrast to \cite{ST-TR}, we use pure self-attention to model skeleton data, and the correlation of all joints in several consecutive frames is calculated at the same time.

\subsection{Skeleton-Based Action Recognition}

Skeleton-based action recognition has been widely studied for decades. Previously, skeleton-based motion modeling methods mainly used 3D information of joints to design handcrafted features \cite{handcrafted_2014,handcrafted_2017}. With the breakthrough of high-performance computing and deep learning technology, deep learning shows excellent ability to extract features.
At present, the deep learning methods for skeleton-based action recognition are mainly divided into three categories: 
(1) The RNN-based methods \cite{RNN_GRU_2017,RNN_LSTM_20171,RNN_LSTM_20172,HSR-TSL} is based on the natural time properties of the skeleton sequence, and then modeled by Long Short Term Memory (LSTM), Gated Recurrent Unit (GRU), etc.; 
(2) The CNN-based methods \cite{CNN_20171,CNN_20172,CNN_2020,STVIM} usually convert the skeleton sequence into the pseudo-images using specific transformation rules, and model it with efficient image classification networks. The CNN-based methods usually combined with RNN-based methods to model the temporal information of skeleton sequence. 
(3) The GCN-based \cite{ST-GCN,Two-Stream,MS-G3D} methods utilize the natural topology and regular sequence of skeleton data in space and time to encode the skeleton into spatio-temporal graph, and use graph convolution network to model it. 
Unlike the above-mentioned methods that represent skeleton data as images or graph, we directly use self-attention to model the skeletal data.

\subsection{Context Aware-based Methods}

The context aware-based methods \cite{Two-Stream,CA-GCN,TEM,TE-GCN,MS-G3D} are designed to extract the features of spatio-temporal non-local joints since they are also related (such as clapping and typing request the cooperation of both hands).
\cite{Two-Stream} superimposes an adaptive matrix on the fixed adjacency matrix to learn the non-local relationship between joints. This method alleviates the limitation caused by the fixed graph topology of the existing GCN-based methods. However, the method only connects the same joints between frames, and cannot model the relationship of different joints between frames.
\cite{CA-GCN} proposed a context aware graph convolution model, which uses three different functions of inner product, bi-linear form and trainable relevance score to calculate the correlation between joints, and then embeds it into the graph convolution to enrich the local response of each body joint by using the information of all other joints.
\cite{TEM} focuses on adding connections on adjacent vertices between frames, and extracting additional features based on the extended temporal graph.
Similarly, \cite{MS-G3D} proposed a multi-scale aggregation scheme to separate the importance of nodes in different neighborhoods in order to achieve effective remote modeling. In addition, a unified spatio-temporal graph convolution operator is proposed, which takes the dense cross spatio-temporal edges as skip connections to directly propagate information on the spatio-temporal graph. 
However, the above method can be regarded as an extension of graph topology and cannot ensure that important joints are connected. In this work, self-attention is utilized to model cross spatio-temporal joints and adaptively capture important joints related to human actions.

\section{Method}

In this section, the overall architecture of the proposed method first is summarized. In the follow, the spatio-temporal tuples encoding and positional encoding strategy is introduced. Finally, each component of the spatio-temporal tuples Transformer is described in detail.

\subsection{Overall Architecture}

\begin{figure*}
  \centering
  {\includegraphics[width=\textwidth]{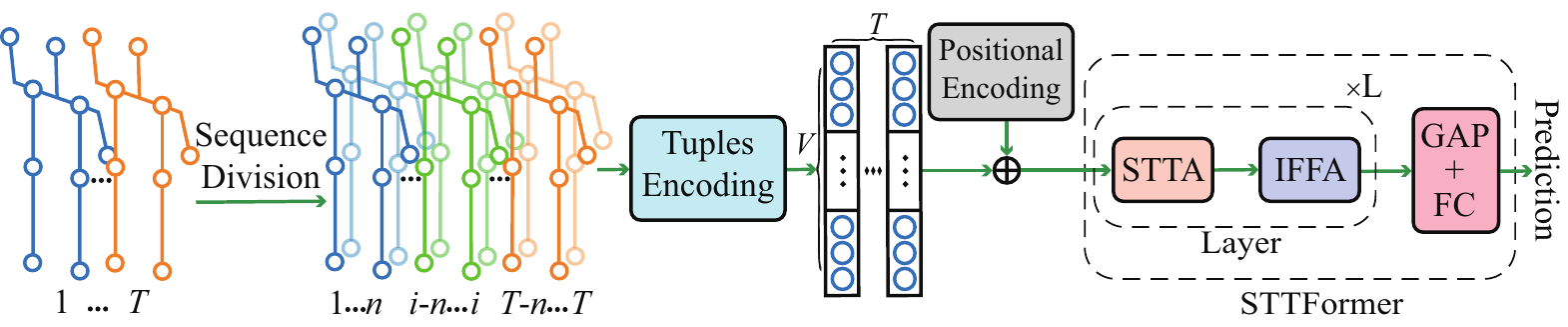}}
  \caption{Illustration of the overall architecture of the proposed model. which consists of two main modules: the spatio-temporal tuples encoding and spatio-temporal tuples Transformer.}
  \label{Model}
\end{figure*}
The overall architecture of our model is shown in Fig. \ref{Model}. The input is a skeleton sequence with $V_0$ joints and $T_0$ frames. We divided the sequence into $T$ parts, each one containing $n$ consecutive frames, a total of $V=n*V_0$ joints. Then a tuple encoding layer is utilized to encode each tuple data. A total of $L$ layers are stacked in spatio-temporal tuples Transformer, and each layer is composed of STTA and IFFA, in which STTA module is used to model the relationship between joints in tuples and IFFA is used to integrate all tuples. Finally, the obtained features are input into a global average pooling layer and a full connection layer to obtain classification scores. In the following sections, the details of each module will be introduced.

\subsection{Spatio-Temporal Tuples Encoding}

\begin{figure}
  \centering
  {\includegraphics[scale=1]{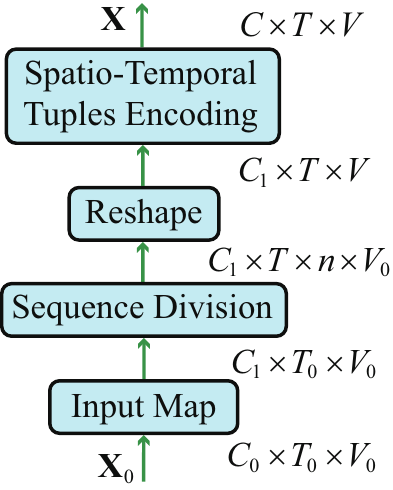}}
  \caption{Illustration of the proposed spatio-temporal tuples encoding module.}
  \label{Encoding}
\end{figure}

To model the relationship between different joints in several consecutive frames, we propose a strategy to encode these joints. The spatio-temporal tuples encoding procedure is illustrated in Fig. \ref{Encoding}.

Firstly, the raw skeleton sequence $\mathbf{X}_0 \in \mathbb{R}^{C_0 \times T_0 \times V_0}$ is fed to a feature mapping layer to expand the input channel to a set number $C_1$. The feature mapping layer is implemented by one convolution layers with BatchNorm and Leaky ReLU function. Subsequently, the skeleton sequence is divided into $T$ non-overlapping tuples:
\begin{equation}\label{X_CTnV0}
  \mathbf{X} = [\mathbf{x}_1, \mathbf{x}_2, \cdots, \mathbf{x}_T], \mathbf{x}_i \in \mathbb{R}^{C_1 \times n \times V_0}
\end{equation} 
where $n$ denotes the number of frames contained in a tuple. In the follow, each tuple is flattened:
\begin{equation}\label{X_CTV}
  \mathbf{X} \in \mathbb{R}^{C_1 \times T \times n \times V_0} \to  \mathbb{R}^{C_1 \times T \times V}
\end{equation} 
where $T=T_0 / n$, $V=n \times V_0$.
Finally, $\mathbf{X}$ is fed to a spatio-temporal tuples encoding layer implemented by one convolution layer with Leaky ReLU function to get the final tuples encoding $\mathbf {X} \in \mathbb{R}^{C \times T \times V}$.

\subsection{Positional Encoding}

The tensor obtained by tuples encoding does not contain the order of joints, and the identity of joints cannot be distinguished, which will reduce the performance of action recognition \cite{DSTANet,IIP-Transformer}, which is also confirmed by the experimental results in Tab \ref{Pos_Embed}. Considering this problem, a position encoding module is used to mark each joint, and the sine and cosine functions with different frequencies are utilized as the encoding functions:
\begin{equation}\label{Pos_Encod}
  \begin{split}
  &PE(p, 2i) = sin(p/10000^{2i/C_{in}}) \\
  &PE(p, 2i+1) = cos(p/10000^{2i/C_{in}})
  \end{split}
\end{equation} 
where $p$ and $i$ denote the position of joint and the dimension of the position encoding vector, respectively.
To model the relationship of all joints in a tuple, it is necessary to distinguish the joints of different consecutive frames, so all joints in a tuple are assigned different IDs.

\subsection{Spatio-Temporal Tuples Transformer} \label{Sec:STT}

\begin{figure}
  \centering
  {\includegraphics[scale=1]{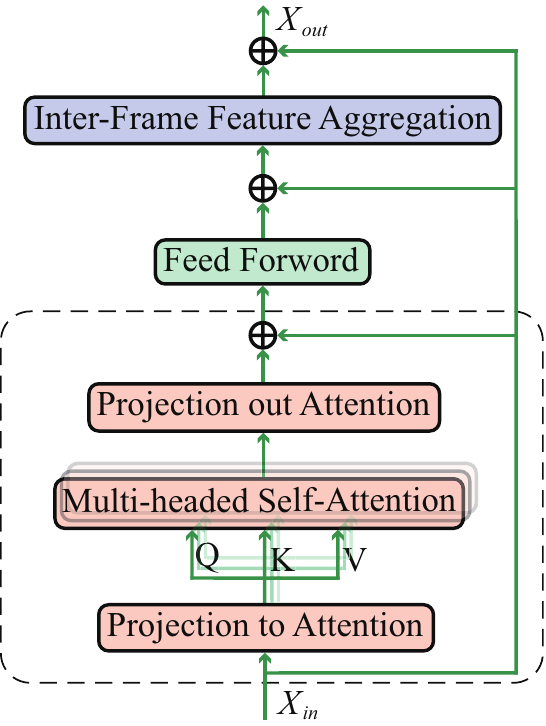}}
  \caption{Illustration of the proposed spatio-temporal tuples Transformer layer, the complete STTFormer is stacked by L such layers.}
  \label{Module}
\end{figure}

As shown in Fig. \ref{Module}, the spatio-temporal tuple Transformer layer includes two main components: spatio-temporal tuple attention and inter-frame feature aggregation, which will be described in detail below.

\subsubsection{Spatio-Temporal Tuples Attention}

The essence of self-attention can be described as the mapping from a query to a series of key and value pairs. After spatio-temporal tuples encoding and positional encoding of skeleton sequence, the multi-headed self-attention mechanism can be used to model the relationship between input tokens. 

Specifically, when calculating self-attention, not only the impact of all other nodes on the node $n_i$, but also the impact of the node $n_i$ on other nodes must be considered. Therefore, the encoded sequence $\mathbf{X}_{in}$ is usually projected into the query $\mathbf{Q}$, key $\mathbf{K}$ and value $\mathbf{V}$. In this work, a convolution layer with $1 \times 1$ kernel size is used to project the encoded sequence $\mathbf{X}_{in}$:
\begin{equation}\label{to_qkv_s}
  \mathbf{Q, K, V} = Conv_{2D(1 \times 1)}(\mathbf{X}_{in})
\end{equation}
Then, the weights can be obtained by calculating the similarity between the query $\mathbf{Q}$ and the transpose of the key $\mathbf{K}$. Like the standard Transformer, the dot-product is simply used as the similarity function. Subsequently, the Tanh function is utilized to normalize the obtained weights. And then the weights and corresponding value $\mathbf{V}$ are weighted and summed to obtain final attention. 
\begin{equation}\label{weights}
  \mathbf{X}_{attn} = Tanh\left(\frac{\mathbf{QK^T}}{\sqrt{C}}\right)\mathbf{V}
\end{equation} 
where $C$ denotes the number of channels of the key $\mathbf{K}$, which can avoid excessive inner product to increase gradients stability during training. Considering the fixed relationship of human joints, learn from DSTANet \cite{DSTANet}, a spatial global regularization is utilized to introduce this relationship. 

To obtain better performance, the multi-headed self-attention mechanism is usually applied, which allows the model to learn related information in different representation subspaces. Specifically, the self-attention operation is performed on multiple groups of $\mathbf{Q, K, V}$ projected by different learnable parameters, and then the multiple groups of attention were concatenated.
\begin{equation}\label{concat_s}
  \mathbf{X}_{Attn} = Concat(\mathbf{X}^{1}_{attn}, \cdots, \mathbf{X}^{h}_{attn})
\end{equation}

In the follow, the obtained $\mathbf{X}_{Attn}$ is projected into an output space by a convolution operation with $1 \times k_1$ kernel size to obtain the result of multi-headed self-attention.

\begin{equation}\label{qkv_out_s}
  \mathbf{X}_{STTA} = Conv_{2D(1 \times k_1)}(\mathbf{X}_{Attn})
\end{equation}
where $k_1$ corresponding to the flattened dimension of spatio-temporal tuples. 

Similar to the transformer, a feed forward layer implemented by 1x1 2D convolution is added to fuse the output.

\subsubsection{Inter-Frame Feature Aggregation}

An action can be regarded as composed of several different sub-actions, such as "long jump" including sub-actions such as "run-up", "take-off" and "landing". In our method, each tuple contains a sub-action, which is obtained by modeling several consecutive $n$ frames using STTA. If the correlation of these sub-actions (such as "run-up", "take-off" and "landing") is constructed, it will help to action recognition and distinguish similar actions (such as high jump and long jump). Therefore, the IFFA operation is proposed to aggregate these sub-actions.
A convolution operation with $k_2 \times 1$ kernel size is used to realize inter-frame feature aggregation in temporal dimension.
\begin{equation}\label{qkv_out_t}
  \mathbf{X}_{IFFA} = Conv_{2D(k_2 \times 1)}(\mathbf{X}_{STTA})
\end{equation}

Finally, the residual connections in Fig. \ref{Module} are also used to stabilize network training. It should be noted that all outputs connected to the rest should be regularized.

\section{Experiments}

In this section, we conducted extensive comparative experiments to evaluate the performance of our method. Firstly, the datasets are described. Then the experimental setup is introduced. In the follow, we conducted extensive ablation studies on NTU RGB+D \cite{NTU60} skeleton sequence data to evaluate the contribution of each parts of our method. Finally, the proposed method is compared with the state-of-the-art methods on NTU RGB+D and NTU RGB+D 120 \cite{NTU120} skeleton sequence data to prove the advantages of our method, and the corresponding analysis is given.

\subsection{Datasets}
 
\vspace{6pt}\noindent\textbf{NTU RGB+D}. NTU RGB+D dataset is a large-scale benchmark for 3D human action recognition captured simultaneously using three Microsoft Kinect V2 sensors. 
The dataset was completed by 40 volunteers and contained 56,000 action sequences in 60 action classes, including 40 daily actions, 9 health-related actions and 11 mutual actions. 
This experiment only uses the skeleton data containing the three-dimensional positions of 25 body joints per frame. 
The dataset is divided into training set and test set by two different standards. 
The Cross-Subject (X-Sub) divides the dataset according to the person ID. the training set and the test set contains 20 subsets respectively. 
The Cross-View (X-View) divides the dataset according to camera ID. The samples collected by cameras 2 and 3 are used for training, and the samples collected by camera 1 are used for testing. It should be noted that the horizontal angles of the three cameras differ by 45° respectively.
 
\vspace{6pt}\noindent\textbf{NTU RGB+D 120}. NTU RGB+D 120 dataset extends NTU RGB+D by adding another 60 classes and another 57,600 samples. The dataset was completed by 106 volunteers and has 114,480 samples and 120 classes in total, including 82 daily actions, 12 health-related actions and 26 mutual actions. 
Like NTU RGB+D, the dataset is also divided by two different standards.
For Cross-Subject (X-Sub), the 106 subjects are split into training and testing groups. Each group consists of 53 subjects.
For Cross setup (X-Set) takes samples with even collection setup IDs as the training set and samples with odd setup IDs as the test set.

\subsection{Experimental Setting}
 
All experiments were performed on 2 GTX 3090 GPUs. All skeleton sequences are padded to 120 frames by replaying the actions. Our model is trained using a Stochastic Gradient Descent (SGD) optimizer with Nesterov momentum 0.9 and weight decay 0.0005, using cross entropy as the loss function. The training epoch is set to 90, the initial learning rate is 0.1, and it is adjusted to one-tenth at 60 and 80 epochs respectively. The batch size is 64. 
Each tuple contains 6 consecutive frames, that is, $n=6$. The number of spatio-temporal self-attention layers is set to 8, and the output channels are 64, 64, 128, 128, 256, 256, 256, and 256, respectively.

\subsection{Ablation Studies}
 
In this section, the effectiveness of the proposed method is investigated on NTU RGB+D Skeleton dataset. For fair comparison, other settings are the same except for the explored object.

\subsubsection{Ablation Studies for Position Encoding and IFFA}

The effect of the position encoding and the IFFA module are investigated. As shown in Tab. \ref{Pos_Embed}, the accuracy of the STTFormer model without position encoding is lower lower than that of complete model, it shows that position encoding can significantly improve performance. The main reason is that different spatio-temporal joints play different roles in an action, and reasonable use of this sequence information will effectively improve performance.
\begin{table}
  \centering
  \caption{Ablation studies for position encoding and IFFA on the NTU RGB+D Skeleton dataset in joint mode. PE denotes position encoding.}
  \begin{tabular}{lcc}
    \toprule
    Method            & X-Sub (\%) & X-View (\%)  \\ 
    \midrule
    STTFormer without PE   & 89.3     & 91.8     \\
    \midrule
    STTFormer without IFFA & 84.5     & 88.1     \\
    \midrule
    STTFormer              & 89.9     & 94.3     \\
    \bottomrule
  \end{tabular}
  \label{Pos_Embed}
\end{table}

Tab. \ref{Pos_Embed} also shows the effect of the IFFA module. It can be found that removing the IFFA module will seriously reduce the performance. The main reason is that the module can effectively model the relationship between sub-actions, which is conducive to distinguish similar actions, thereby improving the performance of the model.

\subsubsection{Ablation Studies for STTA}

In order to verify the effectiveness of our STTA module, a set of comparative experiments are constructed. The number of consecutive frames $n$ contained in each tuple is set to 1 and 6, respectively. $n=1$ means that only the relationship between intra-frame joints is modeled, The scheme is similar to Fig. \ref{Skeleton1}; When $n=6$, it means that the relationship between different joints in 6 consecutive frames is modeled at the same time. To ensure the fairness of the experiment, the kernel size $k_1, k_2$ are set to 1 to eliminate the influence of different convolution kernel sizes.

\begin{table}
  \centering
  \caption{Effect of spatio-temporal tuple attention evaluated on NTU RGB+D Skeleton dataset in joint mode.}
  \begin{tabular}{lcc}
    \toprule
    Method                     & X-Sub (\%) & X-View (\%)  \\
    \midrule
    STTFormer ($n=1; k_1,k_2=1$)    & 82.9    & 86.0     \\
    STTFormer ($n=6; k_1,k_2=1$)    & 86.2    & 91.3     \\
    \bottomrule
  \end{tabular}
  \label{self_attention}
\end{table}

The experimental results are shown in Tab. \ref{self_attention}. It is obvious that the proposed STTA can significantly improve the performance of the model. The main benefit is that the proposed STTA module can not only model the relationship between joints in a frame, but also capture the relationship between different joints in several consecutive frames, so as to improve the performance.

\subsubsection{Effect of Parameters $n$}

\begin{figure}
  \centering
  {\includegraphics[scale=0.5]{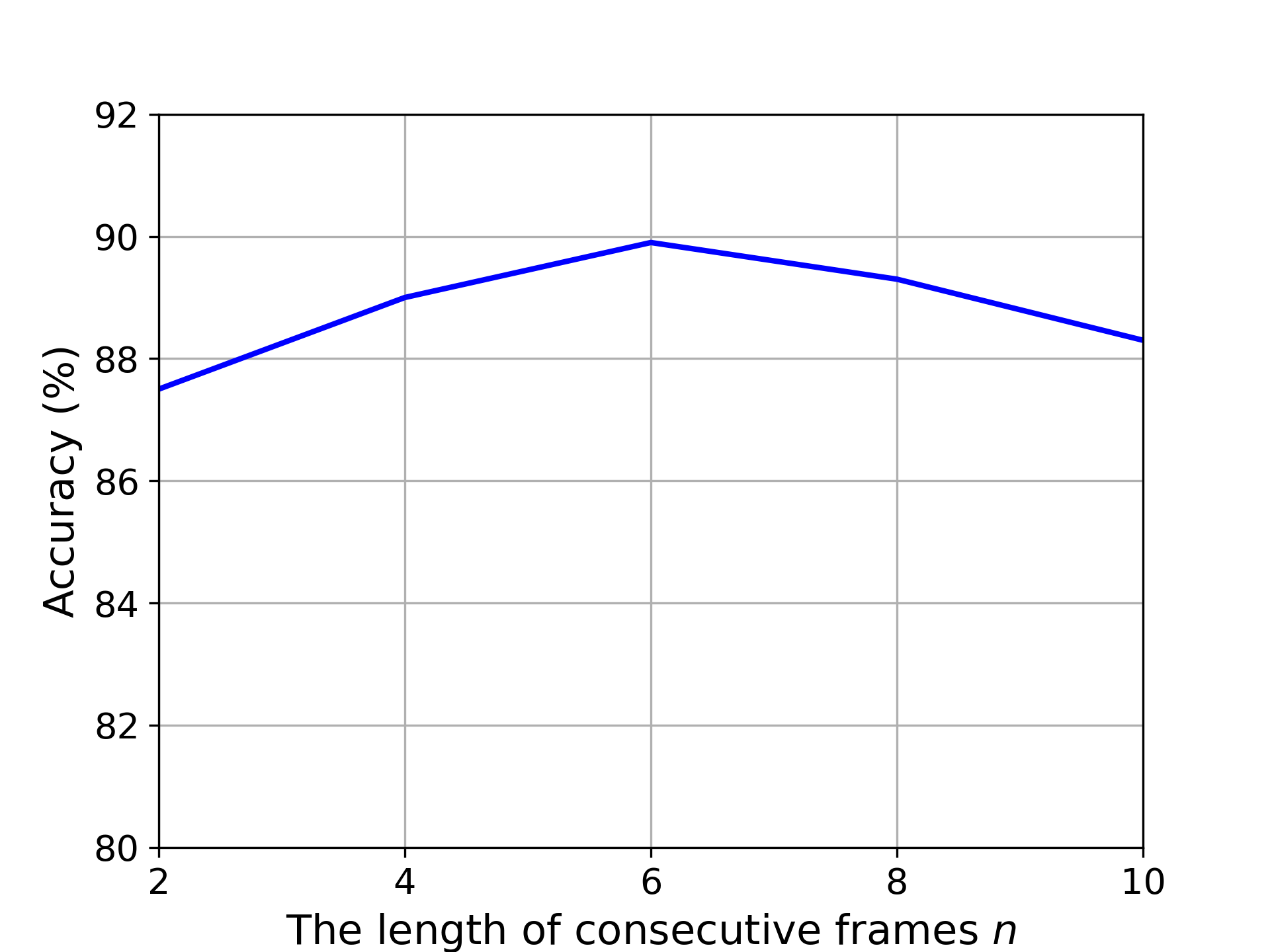}}
  \caption{Effect of the different length $n$ of consecutive frames evaluated on NTU RGB+D Skeleton dataset in joint mode.}
  \label{len_parts}
\end{figure}

The effects of the number of consecutive frames $n$ on our model is explored, as shown in Fig. \ref{len_parts}. 
It can be found that the accuracy of our model is the best when $n=6$, which is consistent with the fact. Because if $n$ is too small, it cannot effectively extract features from the joints of insufficient consecutive frames; And if $n$ is too large, the joint relationship between several consecutive frames is too complex, and the correlation between the first and last frames of each part is low.

\subsubsection{Effect of Multi-Modes Data}

Different patterns of data have different characteristics. Fusing multi-modes data can significantly improve performance. Like most SOTA methods, our model is trained using joint, bone and joint motion modes data respectively, and then average the reasoning outputs of our models to get the final results. The experimental results in Tab. \ref{multi_mode} verify our viewpoint.
\begin{table}
  \centering
  \caption{Ablation studies for the multi-mode data on the NTU RGB+D Skeleton dataset.}
  \begin{tabular}{lcc}
    \toprule
    Method                      & X-Sub (\%)   & X-View (\%)  \\ 
    \midrule
    STTFormer (joint)           & 89.9     & 94.3     \\
    STTFormer (bone)            & 88.8     & 93.3     \\
    STTFormer (joint motion)    & 87.0     & 93.6     \\
    \midrule
    Fusion                      & 92.3     & 96.5     \\
    \bottomrule
  \end{tabular}\label{multi_mode}
\end{table}

\subsection{Comparison with the State-of-the-Art Methods}

The proposed STTFormer method is compared with the state-of-the-art methods on two different datasets: NTU RGB+D, NTU RGB+D 120 Skeleton. Tab. \ref{Compare_SOTA} shows the comparison of recognition accuracy. The comparison methods include CNN-based, RNN-based, GCN-based and Transformer-based methods.

\begin{table} 
    \centering
    \caption{Recognition accuracy comparison against state-of-the-art methods on NTU RGB+D and NTU RGB+D 120 Skeleton dataset.}
    \begin{tabular}{lccccc}
    \toprule
    \multirow{2}{*}{Methods}     &\multicolumn{2}{c}{NTU RGB+D}     &\multicolumn{2}{c}{NTU RGB+D 120}\\
    \cmidrule(r){2-3}\cmidrule(r){4-5}
                                 & X-Sub (\%) & X-View (\%) & X-Sub (\%) & X-Set (\%) \\
    \midrule
    MTCNN\cite{MTCNN}            & 81.1          & 87.4          & 61.2           & 63.3   \\
    IndRNN\cite{IndRNN}          & 81.8          & 88.0          & -              & -      \\
    HCN\cite{HCN}                & 86.5          & 91.1          & -              & -      \\
    \midrule
    ST-GCN\cite{ST-GCN}          & 81.5          & 88.3          & -              & -      \\
    2s-AGCN\cite{2s-AGCN}        & 88.5          & 95.1          & 82.9           & 84.9   \\
    DGNN\cite{DGNN}              & 89.9          & 96.1          & -              & -      \\
    Shift-GCN\cite{Shift-GCN}    & 90.7          & 96.5          & 85.9           & 87.6   \\
    Dynamic-GCN\cite{Dynamic-GCN}& 91.5          & 96.0          & 85.9           & 87.6   \\
    MS-G3D\cite{MS-G3D}          & 91.5          & 96.2          & 86.9           & 88.4   \\
    MST-GCN\cite{MST-GCN}        & 91.5          & \textbf{96.6} & 87.5           & 88.8   \\
    \midrule
    ST-TR\cite{ST-TR}            & 89.9          & 96.1          & 82.7           & 84.7   \\
    DSTA-Net\cite{DSTANet}       & 91.5          & 96.4          & 86.6           & 89.0   \\
    \midrule
    STTFormer(Ours)              & \textbf{92.3} & 96.5          & \textbf{88.3}  & \textbf{89.2}    \\
    \bottomrule
    \end{tabular}\label{Compare_SOTA}
\end{table}

Compared with the CNN and RNN-based methods \cite{MTCNN,IndRNN,HCN}, our proposed STTFormer has significant advantages. The main reason for the poor performance of the CNN and RNN-based methods is that they cannot fully utilize the information of skeleton data. In contrast, the GCN based methods can effectively use the topology of skeleton data and and has better recognition performance.

Compared with the GCN-based methods \cite{ST-GCN,2s-AGCN,DGNN,Shift-GCN,Dynamic-GCN,MS-G3D,MST-GCN}, our method has certain advantages. The reason is that some actions depend on non-local joint interaction, and the STTFormer can fully utilize the interaction information.

Compared with the Transformer-based methods \cite{ST-TR,DSTANet}, our method also achieves state-of-the-art performance. The main benefit is that the STTFormer makes effective use of the correlation of non-local joints between frames.

\section{Conclusion}
In this work, we propose a novel spatio-temporal tuples Transformer (STTFormer) method for skeleton-based action recognition. Our method consists of two modules: STTA and IFFA. The STTA module is used to effectively capture the relationship between different joints in continuous frames, and the IFFA module is used to aggregate the characteristics of several sub-action segments. Ablation studies show the effectiveness of the proposed method. On two large-scale datasets NTU RGB+D and NTU RGB+D 120 Skeleton, the proposed STTFormer achieves better performance with the existing state-of-the-art methods.

 \end{document}